\documentclass[10pt,twocolumn,letterpaper]{article}

\usepackage[pagenumbers]{cvpr} 

\usepackage{graphicx}
\usepackage{amsmath}
\usepackage{amssymb}
\usepackage{booktabs}
\usepackage{multirow}
\usepackage{tipa}
\usepackage{algorithm}
\usepackage{algorithmic}

\usepackage[pagebackref,breaklinks,colorlinks]{hyperref}

\usepackage[capitalize]{cleveref}
\crefname{section}{Sec.}{Secs.}
\Crefname{section}{Section}{Sections}
\Crefname{table}{Table}{Tables}
\crefname{table}{Tab.}{Tabs.}

\begin{document}

\title{Emage: Non-Autoregressive Text-to-Image Generation}

\author{Zhangyin Feng, Runyi Hu, Liangxin Liu, Fan Zhang, Duyu Tang \\
Yong Dai, Xiaocheng Feng, Jiwei Li, Bing Qin,  Shuming Shi
}
\maketitle

\begin{abstract}
Autoregressive and diffusion models drive the recent breakthroughs on text-to-image generation.
Despite their huge success of generating high-realistic images, a common shortcoming of these models is their high inference latency  \textemdash \  autoregressive models run more than a thousand times successively to produce image tokens and diffusion models convert Gaussian noise into images with many hundreds of denoising steps.
In this work, we explore non-autoregressive text-to-image models that efficiently generate hundreds of image tokens in parallel.
We develop many model variations with different learning and inference strategies, initialized text encoders, etc.
Compared with autoregressive baselines that needs to run one thousand times, our model only runs 16 times to generate images of competitive quality with an order of magnitude lower inference latency.
Our non-autoregressive model with 346M parameters
generates an image of 256$\times$256 with about one second on one V100 GPU.
\end{abstract}

\section{Introduction}
Text-to-image generation \cite{mansimov2015generating,nguyen2017plug,xu2018attngan,li2019object,gu2022vector,rombach2022high,feng2022ernie,balaji2022ediffi}, which is to synthesize images from a natural language description, receives growing attention from the research community and the public.
Recently, there has been remarkable progress in text-to-image generation,  owing to hundreds of millions of text-image pairs available from the internet, large neural network models with billions of parameters and infrastructure that supports distributed model training with hundreds of GPUs.
It is reported that text-to-image techniques have been adopted by more than 3000 artists from more than 118 countries in their creative workflows\footnote{\url{https://openai.com/blog/dall-e-2-extending-creativity/}}.

Recent surge of interest in text-to-image generation is due to the impressive ability to deeply understand language  and produce high-fidelity photorealistic images.
Leading approaches could be largely grouped into two model paradigms: autoregressive models and diffusion models.
Autoregressive models (e.g., DALL-E \cite{dalle} and Parti \cite{parti}) typically use Transformer \cite{vaswani2017attention} to convert 
a sequence of words to a sequence of image tokens.
The autoregressive property of these models is that the generation of an image token depends on the previously generated outputs. Therefore, generating the 1024 image tokens of an image, which is a commonly used configuration \cite{dalle,parti}, needs 1024 forward passes of the Transformer decoder.
Diffusion models (e.g., DALL-E 2 \cite{dalle-2} and Imagen \cite{imagen}) 
use iterative denoising steps to convert Gaussian noises to images from a learned data distribution. They usually require more than a hundred encoder-decoder steps to produce an image.
Given the high latency of autoregressive and diffusion models, it is essential to develop efficient models to dramatically speed up the inference process.

\begin{table}[t]
\begin{center}
\begin{tabular}{lcc}
\toprule
\textbf{Approach} && \textbf{Model Type}\\
\midrule
TReCS~\cite{koh2021text}  && GAN   \\
XMC-GAN~\cite{zhang2021cross}  && GAN   \\
DALL-E~\cite{dalle}  && Autoregressive   \\
CogView~\cite{ding2021cogview}  && Autoregressive   \\
CogView2~\cite{cogview2}  && Autoregressive  \\
GLIDE~\cite{nichol2021glide}  && Diffusion \\
Make-A-Scene~\cite{gafni2022make}  && Autoregressive \\
DALL-E 2~\cite{dalle-2}  && Diffusion  \\
Imagen~\cite{imagen}  && Diffusion   \\
Parti~\cite{parti}  &&  Autoregressive  \\
\midrule
\textbf{Emage (this work)} && \textbf{Non-Autoregressive}\\
\bottomrule
\end{tabular}
\caption{Comparison with previous works.
}
\label{table:coco_results}
\end{center}
\end{table}

In this work, we explore non-autoregressive text-to-image models\footnote{We name these non-autoregressive models as \textbf{Emage} (\textbf{E}fficient text-to-i\textbf{ma}ge \textbf{ge}neration). 
}, which are capable of generating hundreds of image tokens in parallel.
Inspired by the development of non-autoregressive model in natural language processing \cite{gu2017non,cmlm,smart,gu2022non}, we develop many model variations, including fully non-autoregressive models that produce all image tokens in one forward pass and iterative non-autoregressive models that iteratively improve the decoding results.
We train these non-autoregressive models on text-image pairs collected from the Internet and have the following findings.
\begin{itemize}
    \item The learning of fully non-autoregressive model, which produces all vision tokens in one forward pass, hardly converges. A successful recipe is to iteratively improve decoding results in a mask-predict manner.
    \item Among three iterative non-autoregressive models, the best one simultaneously makes predictions based on partially-observed image tokens and revises mistakes made in earlier iterations.
    \item Our iterative non-autoregressive models are capable of generating 1024 image tokens with 16 iterations, an order of magnitude faster than autoregressive models.
\end{itemize}

\section{Related Work}

Machine learning based text-to-image generation approaches could be largely grouped into three categories.
The \textbf{first} category is GAN \cite{goodfellow2014generative} based methods \cite{reed2016generative,zhang2017stackgan}, which jointly train a generator to fool the discriminator and train the discriminator to distinguish between real and fake images. 
The \textbf{second} category is autoregressive models.
They are typically paired with a well-trained discrete variational autoencoder \cite{van2017dvae}, which includes a tokenizer that maps an image to discrete latent variables and a de-tokenizer that reconstructs an image from discrete latent variables. 
With the tokenizer and de-tokenizer, text-to-image generation could be readily handled with autoregressive sequence generation using Transformer \cite{vaswani2017attention}.
DALLE \cite{dalle} uses a decoder-only architecture like GPT \cite{radford2019language} and Parti \cite{parti} adopts encoder-decoder  architecture with a pretrained text encoder.
Note that a commonly used setting, which is used in both DALLE and Parti, is that the length of image tokens of a $256 \times 256$ image is 1024.
Since autoregressive models generate a target token depending on the previously generated tokens, their latency is very high in the inference stage. The goal of this work is to significantly accelerate the inference speed by developing non-autoregressive models, so that a thousand image tokens could be generated in parallel with one forward pass.
The \textbf{third} category is diffusion models. They regard image generation as an iterative refinement process, in which the beginning and the end of the refinement process are a Gaussian noise and a real image, respectively. 
DALLE-2 \cite{dalle-2} uses CLIP \cite{clip} as the text encoder and builds a prior model to predict image embeddings, which are further fed to a U-Net architecture \cite{nichol2021improved} to conduct diffusion process.
Compared to DALLE-2, Imagen \cite{imagen} does not need an additional prior model and uses T5 \cite{raffel2020exploring} as the text encoder. 
From the perspective of parallel decoding, each denoising step of a diffusion model is achieved with one forward step, which is also efficient. However, they typically require more than one hundred denoising steps to gradually refine the image. 
On the contrary, our non-autoregressive models needs only a dozen of steps to produce an image (Section \ref{sec:effect-iterative-steps} given a detailed analysis).

 \begin{figure}[b]
    \centering
    \begin{subfigure}{\linewidth}
        \centering
        \includegraphics[width=\linewidth]{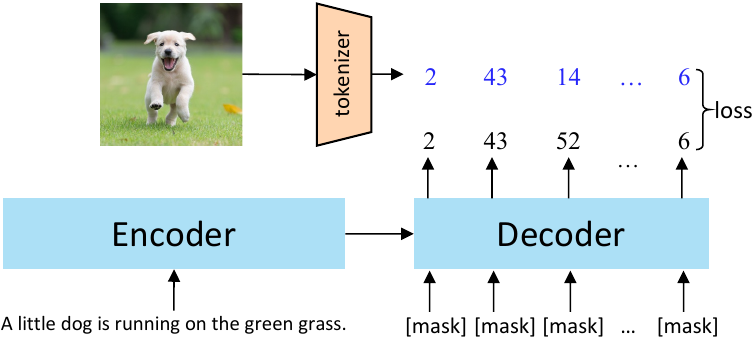}
        \caption{}
    \end{subfigure}
    \begin{subfigure}{\linewidth}
        \centering
        \includegraphics[width=\linewidth]{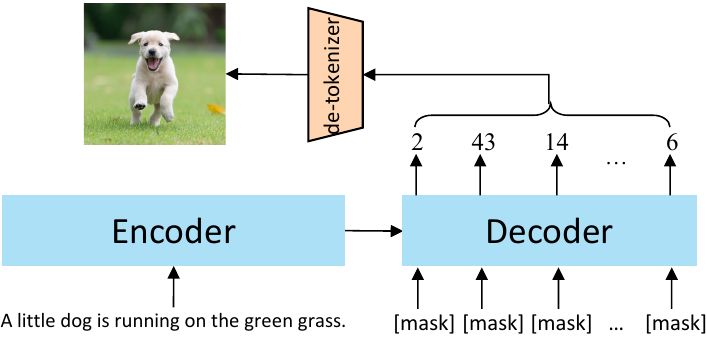}
        \caption{}
    \end{subfigure}
    \caption{Illustration of the training (a) and inference (b) stages of  the fully non-autoregressive text-to-image generation model. }
    \label{fig:method-overall}
\end{figure}

\begin{figure*}
  \centering
    \begin{subfigure}{0.49\linewidth}
    \centering
    \includegraphics[width=0.85\linewidth]{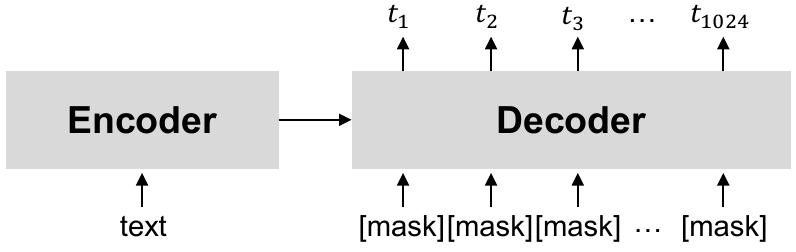}
    \caption{Fully Non-Autoregressive Model}
    
  \end{subfigure}
\vspace{2em}
\hfill
  \begin{subfigure}{0.49\linewidth}
  \centering
    \includegraphics[width=0.85\linewidth]{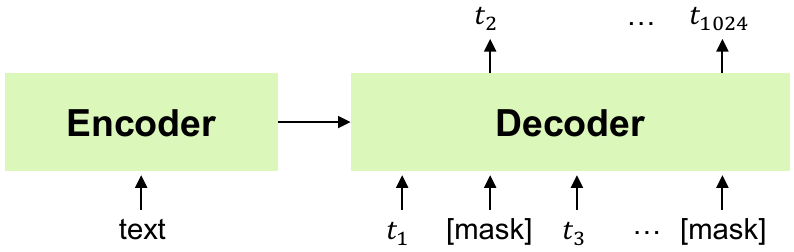}
    \caption{Iterative Non-Autoregressive Model \#1}
    
  \end{subfigure}
  \hfill
  \begin{subfigure}{0.49\linewidth}
  \centering
    \includegraphics[width=0.85\linewidth]{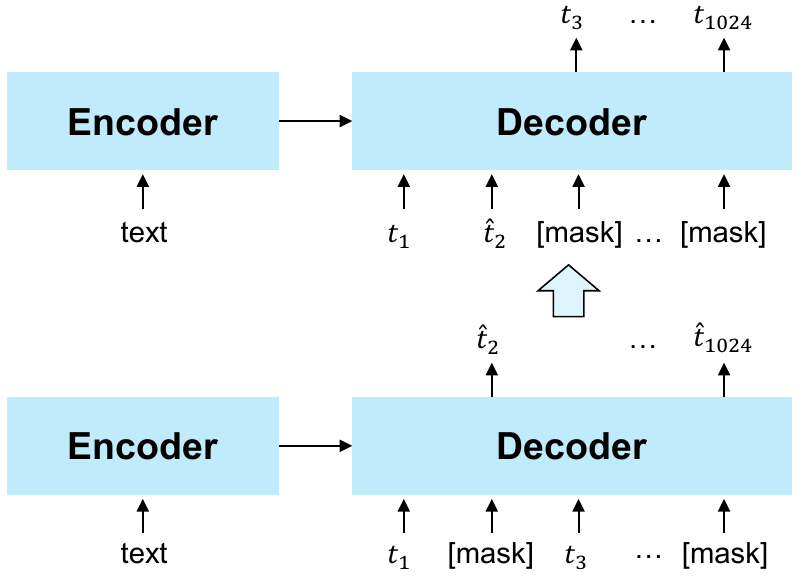}
    \caption{Iterative Non-Autoregressive Model \#2}
    
  \end{subfigure}
  \hfill
  \begin{subfigure}{0.49\linewidth}
  \centering
    \includegraphics[width=0.85\linewidth]{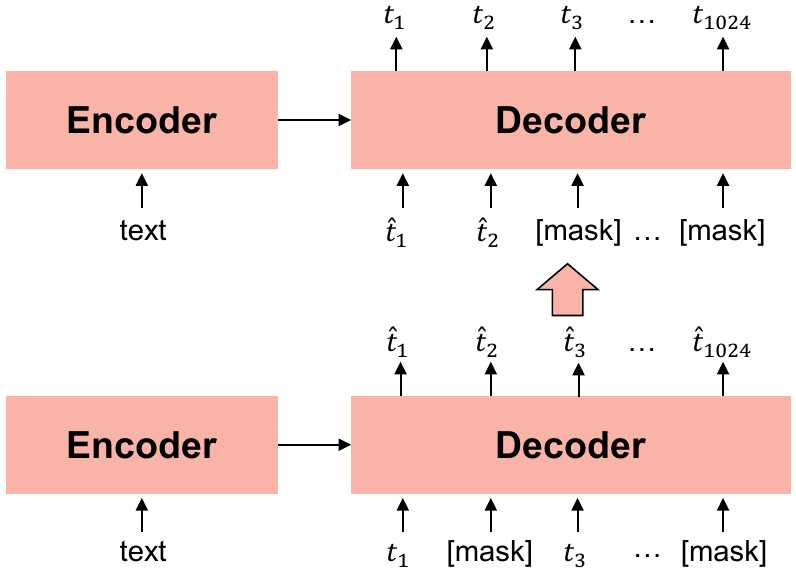}
    \caption{Iterative Non-Autoregressive Model \#3}
    
  \end{subfigure}
  \caption{Training of non-autoregressive text-to-image generation models. Fig (a) is a fully non-autoregressive model that generates all tokens from special tokens in one forward pass. Fig (b) is an iterative non-autoregressive model, which considers the dependency among image tokens with partially-observed image tokens as contexts. Fig (c) extends (b) by mixing predicted image tokens as contexts to mitigate the discrepancy between training and inference stages. Fig (d) not only predicts for unknown positions but revises previously generated image tokens, resulting in globally consistent results. Fig (c) and Fig (d) both need two forward passes in the training process.}
  \label{fig:intro-method}
\end{figure*}

There are attempts on improving the efficiency of autoregressive models.
Cogview 2 \cite{cogview2} uses autoregressive model to generate low-resolution images and use hierarchical decoding to generate local windows in parallel when conducting upsampling. 
Draft-and-revise \cite{lee2022draft} tokenizes an image to a 3D map of  $H \times W \times D$. It produces vectors for $H \times W$ dimensions in parallel and generates $D$ tokens for each of $H \times W$ vector autoregressively. It conducts revisions twice separately by randomly masking a portion of tokens and predicting for those masked positions.
Different from these methods, our non-autoregressive models produces all image tokens in parallel, and if our model (i.e., Figure \ref{fig:intro-method} (d)) does revision, it simultaneously revises the previously generated tokens and generates new tokens simultaneously in each iteration.
Although MaskGit \cite{chang2022maskgit} and M6-UFC \cite{zhang2021m6} also adopt non-autoregressive models for image generation, they either do not study text-conditioned image generation or only experiment with limited scenarios like fashionable clothing and human faces.
In addition, they do not support revising mistakes made in the generation loop.

\section{Non-Autoregressive Models}\label{sec:method}
In this section, we first describe a fully non-autoregressive model and then present three iterative non-autoregressive models.

\subsection{Fully Non-Autoregressive Model}
Our exploration starts from developing a fully non-autoregressive model, which is the most basic and intuitive non-autoregressive solution.
We take this fully non-autoregressive model as an example to illustrate the full picture of the model training and inference process.

The overall framework consists of two components.
The \textbf{first} component is an image vector quantised variational auto-encoder (VQVAE) \cite{van2017dvae}. It includes (1) an image tokenizer that maps an image to a sequence of discrete image tokens, which could be considered as a sequence of words from a special language of image, and (2) an image de-tokenizer that reconstructs an image from image tokens. 
We follow existing autoregressive models \cite{dalle,parti} to use a separately trained VQVAE and transform the problem of text-to-image generation to text-to-image-token generation.
The \textbf{second} component, which is the focus of this work, is 
a non-autoregressive model that maps text description to image tokens.

Let's denote the input text sequence as $ X $
and denote the sequence of target image tokens as $ Y = \{ y_1, y_2, ..., y_n\}$. In our experiments, we set $n$ as 1024 and each image token $y_i$ is an element from a vocabulary of 8192.
Unlike autoregressive models that generate an image token depending on all previously generated tokens, this fully non-autoregressive model is expected to generate all the image tokens in parallel, in one forward pass.
The probability of generating the output sequence $Y$ is calculated as follows,
\begin{eqnarray}
	P(Y|X) = \prod_{i=1}^{n}p(y_i|X ) \nonumber
\end{eqnarray}
where $p(y_i|X)$ indicates the probability of generating image token $y_i$ at the $i$-th position.
Figure~\ref{fig:method-overall} illustrates the training and inference process of this fully non-autoregressive model.
In the inference stage, image tokens are sampled from model's output probability.
The model is trained with the standard cross-entropy loss.

Even though this fully non-autoregressive model has an extremely low inference latency, it totally neglects the dependency among image tokens, which is important for decoding a long sequence of more than a thousand tokens. 
In our experiment, we find that the training of this model hardly converges (see section \ref{learning_curves} for details). 
In the following subsections, we describe three iterative non-autoregressive models, all of which consider the dependency of image tokens and take a few iterations (e.g., 16) to gradually complete the generation of an image given partially-observed image tokens.

\subsection{Iterative Non-Autoregressive Model \#1}\label{sec:iter-model-1}
In this part, we present the first iterative non-autoregressive model.
The basic idea is that the generation of an image, equivalent to the generation of 1024 image tokens, is conducted in a few iterations. 
Let's denote the number of iterations as $T$ (e.g., 16).
For simplicity, we present with linear masking strategy here.
In each iteration, $\frac{1024}{T}$ image tokens are generated in parallel, conditioned on all the image tokens generated in the earlier iterations. 
As the generation process moves forward, the number of conditioned image tokens is increasing and all the image tokens are generated when the last iteration is completed.
This approach fundamentally differs from autoregressive model because the latter adopts token-by-token generation and requires 1024 decoding steps to produce 1024 image tokens. 
In our approach, if $T$ is set to 16, in each iteration 64 image tokens are generated simultaneously.

In the training stage, we randomly mask $k$ image tokens, where $k$ is sampled from a uniform distribution ($k \sim \text{Uniform} (0, n]$) and $n$ is the number of image tokens (e.g., 1024). 
The model learns to predict the original image tokens of the masked positions, conditioned on the text description $X$ and partially-observed image tokens which are not masked out.

\begin{algorithm}[t]
    \caption{Iterative Non-Auto Model \#1 Inference}
    \label{alg:model1_inference}
    \begin{algorithmic}[1]
    \REQUIRE text sequence ${X}$, model  $\theta$, number of iterative  $T$, target length $n$, masking ratio function $\tau()$
    \STATE Set $Y_{obs}$ as a sequence of \texttt{[mask]} of length $n$,  $t = 1$, set $P_{obs}$ as a sequence of zero of length $n$
	\WHILE{$t \leq T$}
		\STATE Model $\theta$ takes $X, Y_{obs}$ as input and predicts image tokens for \texttt{[mask]} tokens with probability $P_{obs}$.
		\STATE Update $Y_{obs}$ by keeping top $\lfloor \tau(\frac{t}{T}) \times n \rfloor$ image tokens ranked by the probabilities $P_{obs}$. 
		\STATE Update the probabilities $P_{obs}$ by setting the values of these $\lfloor \tau(\frac{t}{T}) \times n \rfloor$ tokens as one.
		\STATE $ t = t + 1$
	\ENDWHILE
	\RETURN $Y_{obs}$
    \end{algorithmic}
\end{algorithm}

In the inference stage, we adopt iterative decoding.
Algorithm~\ref{alg:model1_inference} summarizes the inference process.
Let's denote the number of iterations as $T$.
Before the generation starts, we have a sequence of \texttt{[mask]} tokens of the target length $n$ (e.g., 1024), denoted as $Y_{obs}$. 
The whole generation process can be viewed as gradually updating $Y_{obs}$ through filling in the \texttt{[mask]} tokens. 
As time goes on, the number of \texttt{[mask]} tokens in $Y_{obs}$ decreases and reduces to zero after the $T$-th round is completed.
At each round, we keep top $\lfloor \tau(\frac{t}{T}) \times n \rfloor$ image tokens in total as the updated $Y_{obs}$, ranked by model's output probability $P_{obs}$.
We don't make changes to the previously generated image tokens, which is implemented by setting the probability of past image tokens as 1.0 (line 5).
If we use linear function as the masking strategy $\tau(\cdot)$, as what we have done in the training stage, in each round we will fill in the same amount of \texttt{[mask]} tokens. 
In practice, we find that a concave cosine masking strategy \cite{chang2022maskgit}, which encourages fewer predictions at the beginning and more predictions towards the end, works better so we use cosine function in the inference stage.
At a high level, this model could be viewed as an extension of MaskGit \cite{chang2022maskgit} to text-to-image generation.

\subsection{Iterative Non-Autoregressive Model \#2}
\label{subsec: non-auto model 2}
We present the second iterative non-autoregressive model, which improves the first iterative model by bridging the gap between training and inference  \textemdash \  partially-observed tokens $Y_{obs}$ are model-predicted in the inference stage but come from ground truth during training. 
We address this issue with a simple and effective solution by performing two forward passes in the training process. The inference process is same with the iterative non-autoregressive model \#1.

We briefly describe the training process here. 
In the first forward pass, we randomly mask some positions in $Y$ from a uniform distribution ($\sim \text{Uniform}(0, |Y|]$) and predict image tokens for these masked positions with the model, resulting in $Y_{mix}$ which includes both ground truth and model predicted image tokens.
In the second forward pass, we adopt almost the same training process of iterative model \#1, except to replace $Y$ with $Y_{mix}$.

\begin{algorithm}[t]
    \caption{Iterative Non-Auto Model \#3  Training}
    \label{alg:model3_training}
    \begin{algorithmic}[1]
    \REQUIRE text sequence ${X}$, ground truth image tokens ${Y}$
    \STATE initialize model parameter $\theta$
    \FOR{\textit{training not converges}}
    	\STATE Sample a mask ratio $r$ from Uniform $\sim(0,1]$.
		\STATE Randomly pick $k = \lfloor r \cdot |Y| \rfloor$ positions of $Y$.
		\STATE Obtain $Y_{mask}$ from $Y$ by replacing image tokens at $k$ positions with \texttt{[mask]}.
		\STATE Obtain $Y_{pred}$ by using model $\theta$ to predict image tokens for all positions given $X$ and $Y_{mask}$.
    	\STATE Sample another mask ratio $r$ from Uniform $\sim(0,1]$ and randomly mask $Y_{pred}$, resulting in $Y_{obs}$.
            \STATE Predict image tokens for all positions conditioned on $X$ and $Y_{obs}$.
            \STATE Calculate loss for all positions and update $\theta$.
    \ENDFOR
    \RETURN model $\theta$
    \end{algorithmic}
\end{algorithm}

\subsection{Iterative Non-Autoregressive Model \#3}
We further improve the previously mentioned iterative models by enabling the model to revise the mistakes made in earlier rounds.
We believe that the ability to revise allows the model to produce more globally consistent outputs and is especially valuable in our situation.
The reason might be that the predictions made in earlier rounds are less confident because the decisions are made conditioned on a large number of \texttt{[mask]} tokens.

Algorithm \ref{alg:model3_inference} summarizes the inference process.
Specifically, in each iteration the model parallelly predicts for all the target positions, including both \texttt{[mask]} tokens and previously generated image tokens.
For previously generated tokens, we use newly predicted image tokens to update the values.
These tokens will not be replaced with \texttt{[mask]} in following rounds (line 5).
For the masked positions, in each round we fill in top-ranked image tokens measured by model's output probability. 
After the $t$-th round is completed, there will be $\lfloor \tau(\frac{t}{T}) \times n \rfloor$ image tokens in $Y_{obs}$.

In the training stage, we run two forward passes. 
In the first forward pass, we randomly mask some positions of $Y$ from a uniform distribution ($\sim \text{Uniform}(0, |Y|]$) and use the model to predict tokens for all positions.
We call the sequence of model predicted image tokens as $Y_{pred}$.
In the second forward pass, we repeat the masking process by conducting another sampling from a uniform distribution over $Y_{pred}$, resulting in $Y_{obs}$. 
In this way, $Y_{obs}$ includes both \texttt{[mask]} tokens and wrong image tokens.
The model takes $X$ and $Y_{obs}$ as the input to predict for all target positions and
calculate the cross-entropy loss for all target positions.
In this way, the model learns the ability to predict new tokens and to revise wrong image tokens.
Algorithm \ref{alg:model3_training} summarizes the training process. 

\begin{algorithm}[t]
    \caption{Iterative Non-Auto Model \#3  Inference}
    \label{alg:model3_inference}
    \begin{algorithmic}[1]
    \REQUIRE text sequence ${X}$, model  $\theta$, number of iterative  $T$, target length $n$, masking ratio function $\tau()$
    \STATE Set $Y_{obs}$ as a sequence of \texttt{[mask]} of length $n$, $t = 1$, set $P_{obs}$ as a sequence of zero of length $n$  
   	\WHILE{$t \leq T$}
            \STATE Model $\theta$ takes $X, Y_{obs}$ as the input and predicts image tokens for all positions, including both \texttt{[mask]} tokens and image tokens 
            with probability $P_{obs}$.
            \STATE Update $Y_{obs}$ by 
            keeping top $\lfloor \tau(\frac{t}{T}) \times n \rfloor$ image tokens ranked by the probabilities $P_{obs}$.
            \STATE Update the probabilities $P_{obs}$ by setting the values of these $\lfloor \tau(\frac{t}{T}) \times n \rfloor$ tokens as one.
		\STATE $ t = t + 1$
	\ENDWHILE
	\RETURN  $Y_{obs}$
    \end{algorithmic}
\end{algorithm}

\section{Experiments}

In our experiments, we train autoregressive and non-autoregressive models on the same text-image pairs collected from the Internet. 
We get 280M text-image pairs in total as the training data, including 260M text-image from LAION-400M \cite{schuhmann2021laion}, 5M text-image pairs from CC12M \cite{changpinyo2021conceptual}, 2M text-image pairs from CC3M \cite{sharma2018conceptual} and 13M text-image pairs from YFCC100M \cite{thomee2016yfcc100m}. 
The amount of data is smaller than the original ones because our download successful rate is not perfect and we remove images with no paired caption and use simple rules to filter out low-quality captions.
Since our approach builds on image tokenizer, we explore different public tokenizers including dVAE of \cite{dalle} and VQGAN of \cite{esser2021taming}. Based on our analysis (see section \ref{subsec:img-tokenizer}), we use the Taming Gumbel VQGAN of \cite{esser2021taming}, the last row in Table \ref{table:img-tokenizer-results}, as the final image tokenizer. It represents an image as 1024 tokens, each of which comes from a vocabulary of 8192. We also test different pretrained models including BERT \cite{devlin2018bert} and CLIP \cite{clip} as our text encoder and find that CLIP plays a vital role to the success of non-autoregressive models (see section \ref{subsec:text-enc-select} for detailed results).

We train base non-autoregressive models of 12 Transformer layers with about 250K steps to find the best setting and then train 500K steps for the best setting. 
We set both encoder and decoder with 12 layers and set the both hidden length as 768.
Our models are trained with the following training hyperparameters: learning rate=4.5e-4, batch size=1024, grad clip norm=4.0, AdamW optimizer with $\beta_1$=0.9, $\beta_2$=0.96, weight decay=4.5e-2, cosine learning rate scheduler with warm up ratio=0.02. To stabilize the training process, we use FP32 instead of FP16.

Following existing works, we evaluate models on a subset of randomly sampled 30k images from the MS-COCO \cite{lin2014microsoft} validation set measured with FID \cite{heusel2017gans} in a zero-shot setting (i.e., without using the training set of MS-COCO for model fine-tuning). 
For all the autoregressive and non-autoregressive models, we generate 16 images for each caption. 
For non-autoregressive models, this is achieved by adding gumbel noise to model predicted probabilities when selecting which positions to ``unmask''.
We then rank these candidates by CLIP similarity and select the best one for FID computing.

\subsection{Selection of Image Tokenizer}\label{subsec:img-tokenizer}
In this part, we describe how we select the image VQVAE, which works as the image tokenizer and de-tokenizer of the whole framework, before conducting text-to-image model training.
Specifically, we test the image reconstruction ability of six public models from dVAE of \cite{dalle} and VQGAN of \cite{esser2021taming}. 
Here we tokenize an image to a sequence of tokens and then use de-tokenizer to reconstruct the image.
These models are evaluated by measuring the FID between the predicted image and the ground truth. We test them on the whole validation set of MS-COCO which includes 40K instances.
Note that this setting does not involve text. 

\begin{table}[t]
\begin{center}
\begin{tabular}{c|c|c|c|c}
\toprule
\textbf{Model} & \textbf{Length} & \textbf{Vocab} & \textbf{Code Dim} & \textbf{FID} $\downarrow$ \\
\midrule
dVAE & 1,024 & 8,192 & 128 & 36.23 \\
\hline
\multirow{5}{*}{VQGAN}  & 256 & 1,024 & 256 & 22.35 \\
 & 256 & 16,384 & 256 & 17.73 \\
 & 1,024 & 256 & 4 & 14.60 \\
 & 1,024 & 16,384 & 4 & 14.36 \\
 & 1,024 & 8,192 & 256 & \textbf{12.38} \\
\bottomrule
\end{tabular}
\caption{Results on image tokenization and reconstruction, measured by FID (lower is better) on the MS-COCO validation set. $\downarrow$~means that lower is better.}
\label{table:img-tokenizer-results}
\end{center}
\vspace{-0.2in}
\end{table}

From Table \ref{table:img-tokenizer-results}, we can see that the last row of VQGAN performs better than other tokenizers. Therefore, in the following experiments, we use VQGAN with the above setting as the default image tokenizer and de-tokenizer.

\subsection{Learning Curves: Fully Non-Autoregressive Model Fails to Converge}\label{learning_curves}
\begin{figure}[h]
  \includegraphics[width=1\linewidth]{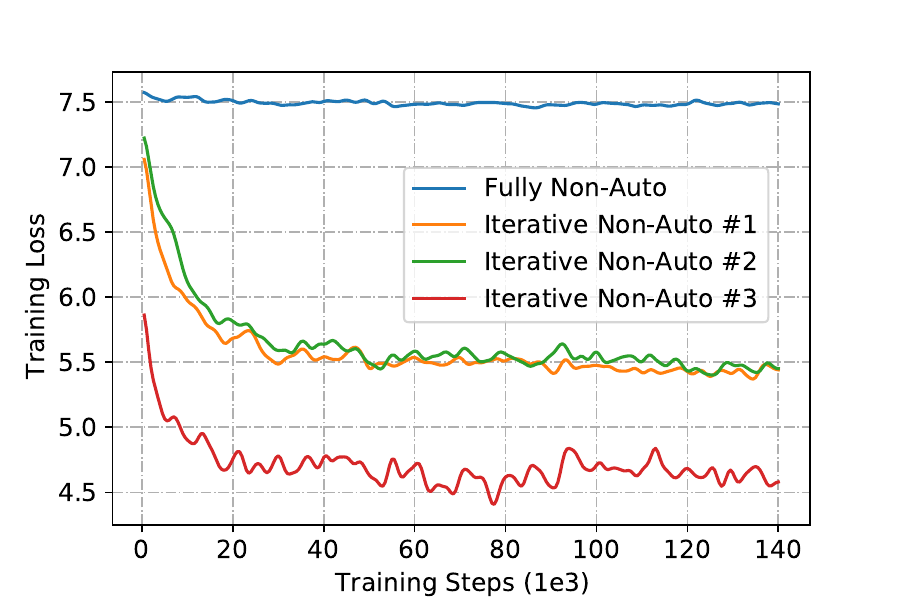}
  \caption{Learning curves for four non-autoregressive models. X-axis and Y-axis represent training steps and the loss, respectively.}
  \label{fig:learning-curves}
\end{figure}

We show the learning curves of four non-autoregressive models in Figure \ref{fig:learning-curves}.
We observe the loss of the fully non-autoregressive model (top line) hardly drops compared with other three iterative non-autoregressive models.
Recap that the fully non-autoregressive model
masks all the image tokens and learns to predict them in one forward pass during training.
We also tried curriculum learning by gradually increasing the \texttt{[mask]} ratio, but did not find improvements.
We use the checkpoint of 140K steps of this fully non-autoregressive model to generate images from caption and find the generated images could only show basic colors and vague outlines. 
The reason might be that generating a sequence of a thousand tokens in parallel, which is significantly longer than the common setting of natural language generation or machine translation, is too challenging 
without any context information from the target side.

\subsection{Comparison Among Base Autoregressive and Non-Autoregressive Models}
We compare the FID and inference latency among our base autoregressive and iterative non-autoregressive models, all of which are trained about 250K steps. All the text encoders are initialized with CLIP large model.
In each iterative, non-autoregressive model generates an image with 16 forward steps. 
The latency of generating an image is computed on a single GPU (V100 with 32G memory) with batch size as one.

\begin{table}[ht]
\begin{center}
\begin{tabular}{l|c|c}
\toprule
\textbf{Model} & \textbf{FID} $\downarrow$ & \textbf{Latency (s)}  \\
\midrule
Autoregressive Model & 19.45 & 48.0 \\
Iterative Non-Auto Model \#1 & 32.59 & 0.95 \\
Iterative Non-Auto Model \#2 & 28.01 & 0.95 \\
Iterative Non-Auto Model \#3 & 24.97 & 0.95 \\
\bottomrule
\end{tabular}
\caption{Comparison among base models (346M) with $\sim$250K training steps by zero-shot FID and latency (seconds) of generating an image on one V100 GPU.}
\label{table:base_comp_results}
\end{center}
\vspace{-0.2in}
\end{table}

Results are given in Table \ref{table:base_comp_results}. 
We can see that the third iterative non-autoregressive model performs best among three iterative models, achieving competitive FIDs compared to the autoregressive counterpart.
More importantly, the latency of three iterative non-autoregressive models are an order of magnitude less than the autoregressive model. 

\subsection{Main Results}
\begin{figure*}[!ht]

    \begin{subfigure}{\linewidth}
    \centering
    \includegraphics[width=0.96\linewidth]{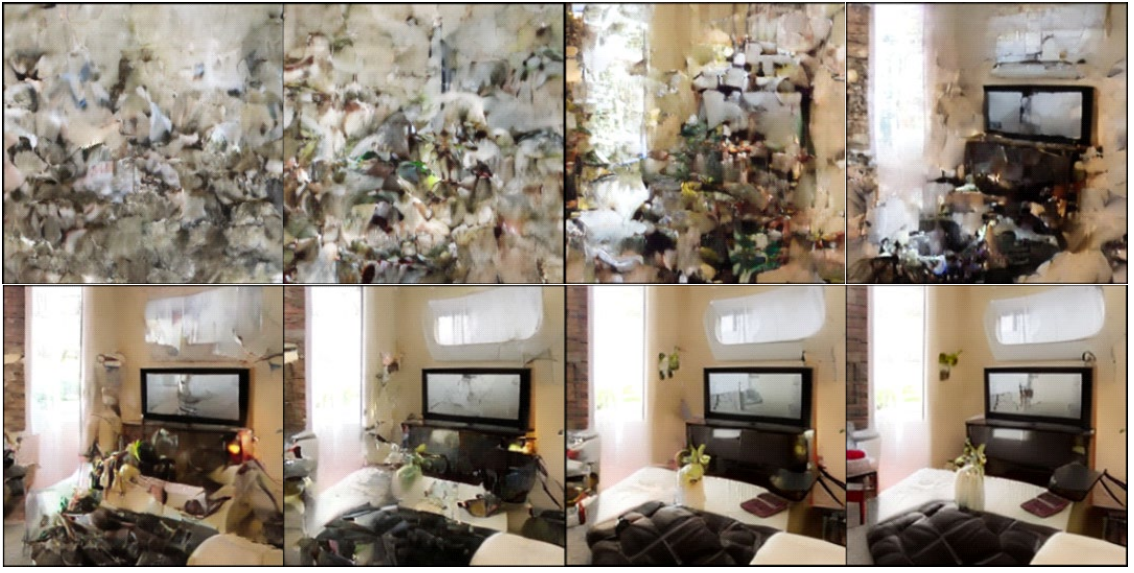}
    \caption{Generated image for ``\textit{a living area with a television and a table.}''.}
  \end{subfigure}
  
      \begin{subfigure}{\linewidth}
    \centering
    \includegraphics[width=0.96\linewidth]{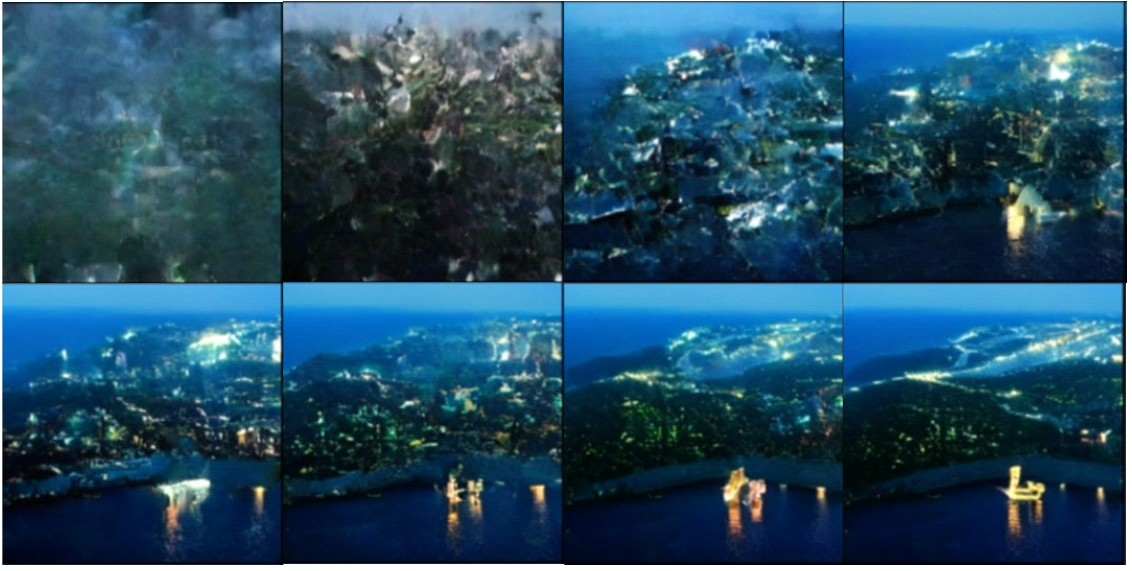}
    \caption{Generated image for ``\textit{aerial view of the beach at night.}''.}
  \end{subfigure}

  \caption{ These eight images are the model outputs at iteration of \{1,3,5,7,9,11,14,16\} from upper left to bottom right. }
  \label{fig:case}
\end{figure*}

We report zero-shot FID on MS-COCO dataset of existing works, our best non-autoregressive model (model~\#3), and our autoregressive baseline which is trained with the same amount of data. 
Models developed by us (second group in Table \ref{table:main-coco_results}) are trained for 500K steps. 
Results show that our best non-autoregressive model achieves competitive FID but is about 50$\times$ faster than the autoregressive baseline of the same model scale.
Case studies are given in Figure \ref{fig:case}. 

\begin{table}[h]
\begin{center}
\begin{tabular}{l|c|c|c}
\toprule
\multirow{2}{*}{\textbf{Approach}} & \multirow{2}{*}{\textbf{Params}}  & \textbf{Zero-shot} & \multirow{2}{*}{\textbf{Latency}}\\
& & \textbf{FID} $\downarrow$ & \\
\midrule
DALL-E~\cite{dalle} & 12B & $\sim$28 & - \\
CogView~\cite{ding2021cogview}  & 4B &  27.1 & - \\
CogView2~\cite{cogview2} & 6B &  24.0 & - \\
GLIDE~\cite{nichol2021glide}  & 5B &  12.24 & - \\
Make-A-Scene~\cite{gafni2022make} & 4B &  11.84& -  \\
DALL-E 2~\cite{dalle-2} & 6.5B & 10.39 & - \\
Imagen~\cite{imagen} & 7.9B & 7.27 & - \\
Parti~\cite{parti} & 20B & 7.23 & - \\
\midrule
Autoregressive  & 346M & 17.00 & 48.0\\
{Emage}  & 346M & 19.74 & 0.95\\
\bottomrule
\end{tabular}
\caption{Model comparison. Latency is the seconds of generating an image
on one V100 GPU.}
\label{table:main-coco_results}
\end{center}
\vspace{-0.2in}
\end{table}

\subsection{Analysis: Importance of CLIP as Text Encoder}\label{subsec:text-enc-select}
Due to the flexibility of our encoder-decoder architecture, we can easily experiment with different types of pretrained text encoder and study their effectiveness for non-autoregressive image generation. 
We take iterative non-autoregressive model \#1  to do this study and try BERT and CLIP with the same number of Transformer layers (i.e., 12) as the text encoder.
BERT \cite{devlin2018bert} is a widely used model for text encoding trained on  text corpus. CLIP \cite{clip} is a popular pretrained model in the multimodal field which is trained with a variety of text-image pairs. 

\begin{table}[ht]
\begin{center}
\begin{tabular}{l|c|c}
\toprule
 & BERT & CLIP \\
\midrule

Autoregressive  & 28.27 & 19.45\\
Iterative Non-Auto Model \#1 & 72.77 & 32.59 \\
\bottomrule
\end{tabular}
\caption{FID of base autoregressive and non-autoregressive models with encoder initialized with BERT and CLIP, respectively.}
\label{table:bert-vs-clip}
\end{center}
\vspace{-0.2in}
\end{table}

FID numbers of both autoregressive model and iterative non-autoregressive model (\#1) trained with about 250K steps are given in Table \ref{table:bert-vs-clip}. We can see that CLIP performs better than BERT in both configurations, demonstrating the effectiveness of text-image pretrained model for our task.
We also observe that
initializing text encoder with CLIP improves the semantic relation between the generated image and the input caption.

\subsection{Analysis: Effects of Inference Steps}
\label{sec:effect-iterative-steps}
We try different inference steps to choose the best number of inference steps that  trades off between inference latency and the quality of generated images. 
Intuitively, setting a small number of inference steps can speed up the decoding process but would hurt the quality of generated images. 
Based on iterative non-autoregressive model \#3, we experiment with different values of \{4, 8, 16, 32\} and find that 16 is a good balance between speed and accuracy, as shown in Table \ref{table:results-iter-num}. 

\begin{table}[ht]
\begin{center}
\begin{tabular}{c|c|c}
\toprule
\textbf{Number of Iteration} & \textbf{FID} $\downarrow$   & \textbf{Latency (s)} \\
\midrule
4 & 71.13 & 0.25 \\
8  & 27.99 & 0.48 \\
16 & 19.74 & 0.95 \\
32 & 18.94 & 1.85 \\
\bottomrule
\end{tabular}
\caption{Results with different inference steps.}
\label{table:results-iter-num}
\end{center}
\vspace{-0.2in}
\end{table}

\subsection{Analysis: Effects of Masking Strategy}\label{sec:mask_strategy}
\begin{figure}[h]
  \includegraphics[width=0.96\linewidth]{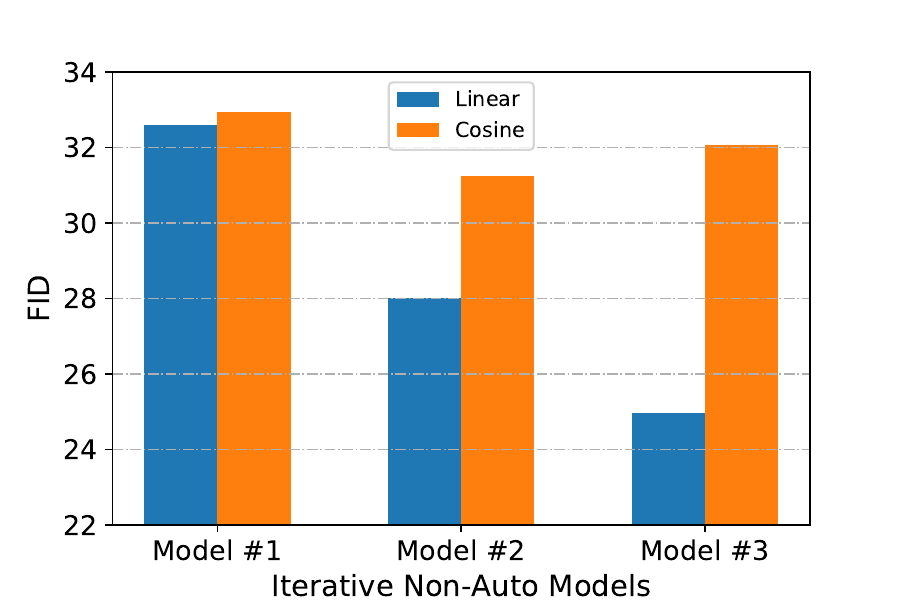}
  \caption{Training with different masking functions.}
  \label{fig:linear_cosine}
  \vspace{-0.2cm}
\end{figure}

As mentioned in Section \ref{sec:method}, masking schedule functions are used in both training and inference of iterative non-autoregressive models. 
Linear function uniformly produces samples with different mask ratios, while cosine function prefers to more masked tokens than linear function.
Cosine function encourages more masked tokens and linear function has no such preference. 
Our observation is that cosine function produces significantly better images than linear function in the inference stage, which is consistent with the observation of MaskGit \cite{chang2022maskgit}.
We adopt cosine as the default setting during inference.
We try training with different masking strategies and show results of models trained with about 250K steps in Figure \ref{fig:linear_cosine}.
We can see that training with linear strategy works better for all iterative non-autoregressive models. 
The reason might be that learning to predict with less and more image tokens as context are both important for our model. 

\subsection{Discussion and Limitations}
We discuss some observations and potential limitations of this work.
First, as revealed in subsection \ref{subsec:text-enc-select}, CLIP embedding plays an important role in the successful training of our non-autoregressive model, compared to a BERT encoder with similar model scale. Results in Imagen \cite{imagen} show that CLIP text encoder of about 120M parameters even performs comparably with a larger text encoder of about 4.6B parameters (T5 XXL). We plan to see whether using a larger text encoder would mitigate this huge performance gap. 
Second, we observe that our model struggles to generate beautiful human face. We plan to scale the model to an order of magnitude large parameters and also improve the data in terms of both image quality, diversity and scale.

\section{Conclusion}
We present non-autoregressive text-to-image generation models dubbed Emage in this work. 
We first observe that fully non-autoregressive model fails to converge in the training process.
Among three iterative non-autoregressive models, we find that the last one which simultaneously predicts new tokens and revises outputs of early rounds performs the best. 
Model analysis suggests the following effective configurations including initializing encoder with CLIP, training with linear masking function and sampling with cosine masking function.
Generating an image with 346M parameters through decoding of 16 rounds takes about one second on one V100 GPU, an order of magnitude faster than autoregressive counterpart.

{\small
\bibliographystyle{ieee_fullname}
\bibliography{egbib}
}

\end{document}